\documentclass[conference]{IEEEtran}
\usepackage{cite}
\usepackage{amsmath,amssymb,amsfonts}
\usepackage{algorithmic}
\usepackage{graphicx}
\usepackage{textcomp}
\usepackage{booktabs,colortbl,multirow,multicol,ctable}
\usepackage{xcolor}
\usepackage{soul}
\usepackage{bm}
\usepackage{amsmath}

\def\BibTeX{{\rm B\kern-.05em{\sc i\kern-.025em b}\kern-.08em
    T\kern-.1667em\lower.7ex\hbox{E}\kern-.125emX}}
\begin{document}
\title{Deep Learning based Covert Attack Identification for Industrial Control Systems\\
}

\author{
    \IEEEauthorblockN{Dan Li\IEEEauthorrefmark{1}, Paritosh Ramanan\IEEEauthorrefmark{1},
    Nagi Gebraeel\IEEEauthorrefmark{1}, and Kamran Paynabar\IEEEauthorrefmark{1}}
    \IEEEauthorblockA{\IEEEauthorrefmark{1}\textit{School of Industrial and Systems Engineering} \\
\textit{Georgia Institute of Technology} Atlanta, USA \\
    \{dli352, paritoshpr\}@gatech.edu} \{nagi.gebraeel, kamran.paynabar\}@isye.gatech.edu
}

\maketitle

\begin{abstract}
Cybersecurity of Industrial Control Systems (ICS) is drawing significant concerns as data communication increasingly leverages wireless networks. A lot of data-driven methods were developed for detecting cyberattacks, but few are focused on distinguishing them from equipment faults. In this paper, we develop a data-driven framework that can be used to detect, diagnose, and localize a type of cyberattack called covert attacks on smart grids. The framework has a hybrid design that combines an autoencoder, a recurrent neural network (RNN) with a Long-Short-Term-Memory (LSTM) layer, and a Deep Neural Network (DNN). This data-driven framework considers the temporal behavior of a generic physical system that extracts features from the time series of the sensor measurements that can be used for detecting covert attacks, distinguishing them from equipment faults, as well as localize the attack/fault. We evaluate the performance of the proposed method through a realistic simulation study on the IEEE 14-bus model as a typical example of ICS. We compare the performance of the proposed method with the traditional model-based method to show its applicability and efficacy.
\end{abstract}


\section{Introduction}
Industrial control systems are collections of complex networks of interactive control and automation systems that are widely used for control and monitoring in chemical, energy, and manufacturing industries \cite{stouffer2011guide}. With the development of sensor technology in the past few decades, modern industrial control systems are generating a massive amount of data in real-time, leveraging the system operations and monitoring to wireless data communication. Although these technologies bring novel insights to the optimization of system operations, they also raise concerns about cybersecurity \cite{desmit2017approach,sridhar2011cyber}. Unlike the common information systems, industrial control systems belong to cyber-physical systems in which the cyber layer is interactive with the physical layer. The computation and the data transformation on the cyber layer enforces the operation and management of the physical layer, which in turn generates data that is fed back to the cyber layer. Because of these operational control interactions, malicious attacks that impact data integrity can cause physical damage to the system and thus be particularly dangerous. For example, the outbreak of the malware \textit{Stuxnet} \cite{kushner2013real} in 2010 draws public attention to data integrity attacks. The event showed such cyberattacks can truly happen and cause severe damage to the critical infrastructures. Therefore, data integrity becomes a very important factor for cyber-physical systems \cite{sridhar2010data}.

 Data integrity attacks often take over the programmable logic controllers (PLC) and smart sensors, and manipulate sensor measurements and/or control actions, while bypassing the traditional detection schemes \cite{mo2014detecting}. For example, \textit{false data injections} \cite{liu2011false} refer to the attacks where the attacker only manipulates the sensor measurements to mislead the state estimation and further alter the control action. \textit{Replay attacks} refer to the attacks where the attacker manipulates the control action and replays the sensor measurements recorded from normal operations to disguise the malicious control. In this paper, we focus on a more sophisticated type of integrity attack called \textit{covert attack} \cite{smith2011decoupled}, where the attacker manipulates the control action and disguises the malicious behavior by replacing the sensor measurements with the simulated data based on the original control input. The underlying assumption includes the following: 1) The attacker has access to manipulate the control actions and sensor measurements; 2) The attacker possesses knowledge of the system behavior so that the simulated data could represent the normal behavior of the system and can be used to mask the malicious manipulation. The covert attack has been proved to be able to be undetectable when the attacker has access to all the sensors and full knowledge of the system \cite{smith2011decoupled}.

Aside from the difficulty in detecting these integrity attacks like covert attacks, there are two challenges in cyberattack identification for ICSs. The first one is the false alarms triggered by natural equipment faults \cite{buczak2015survey}. Although various methods can successfully detect the cyberattack, few of them consider the possibility that a natural equipment fault could trigger false alarms. These false alarms can bring the unnecessary cost of inspections and even system shutdowns. The other challenge is the localization \cite{nudell2015real}. Industrial control systems are networks of multiple subsystems, which serve as the nodes in the network. These nodes may operate individually but are physically connected. Typical examples of ICS are found in power networks, where a control center may be monitoring multiple substations and power generation plants that are geographically distributed, while being physically connected by the transmission lines. Once a cyberattack is detected by the control center, the next step is to determine the location of the attack. Because of the complex interconnectivity among these subsystems, the localization of the attack is nontrivial. For example, a covert attack might impact multiple nodes such that more than one will raise up alarms for detecting an attack. Therefore, a diagnosis method needs to be developed to localize the attack.

In this paper, we demonstrate the application of deep learning towards the detection, diagnosis, and localization of covert attacks. We focus on generic networked industrial control systems and propose a data-driven framework that combines an autoencoder, an RNN, and a DNN. We use the RNN to characterize the system behavior under normal operations. The output of the RNN together with the sensor measurements are fed to a DNN classifier to detect, diagnose, and localize the anomaly. We use the autoencoder to extract features that represent the system status, as well as the spatial correlation among the nodes, in an unsupervised manner. The RNN captures the temporal behavior of the features extracted by the autoencoder, and the DNN helps detect anomalies in the system as well as diagnose whether it is an attack or fault. By considering both the spatial and temporal behavior of the system, this DL framework helps reduce false alarms triggered by natural faults as well as localize the attack by extracting the features that distinguish anomalies at different locations and between attack and faults.
\section{Related Work}
The literature on cyberattack detection for industrial control systems can be divided into two groups: model-based detection and data-driven detection.

Model-based methods rely on the engineering knowledge of the physical rules to establish a parameterized model of the normal sensor measurements \cite{liu2011false,van2015sequential, cardenas2011attacks,huang2018online,li2014quickest}. In \cite{van2015sequential}, \cite{cardenas2011attacks}, and \cite{huang2018online}, the observed sensor measurements are compared with the estimated sensor measurements from these models, and the residuals (i.e., the difference between the observations and model estimations) are monitored to detect the anomalies. The models are often established such that the expectation of the residuals is approximately 0 under normal operations. During monitoring, a large residual means a large discrepancy between the model estimation and the observations, which indicates anomalous behavior of the system. The $\chi^2$ detector, for example, tests the sum of squared residuals (SSR) and triggers alarms whenever the SSR is above the threshold defined by a $\chi^2$ distribution. The drawback of the model-based methods is the complexity in establishing an accurate physical model, especially when the system consists of multiple subsystems with complex connections and correlations, both spatially and temporally. For complex systems such as power systems, the system model is usually represented by the steady-state power flow equations, which takes the observations as independent inputs and does not consider the system dynamics over time.

Compared to model-based methods, data-driven methods are more flexible. A number of applications of machine learning techniques have been developed for detecting cyberattacks. However, most of them focus on designing intrusion detection systems based on the network traffic data \cite{yilmaz2017mitigating,terai2017cyber,caselli2016specification}. If such intrusion detection systems fail, there is a lack of backup detection scheme that detects cyberattacks based on the physical process. On the other hand, unlike the behavior of network traffic, the physical process has a relatively stable and consistent dynamic behavior that does not change over time. Therefore, a recurrent neural network can be used to characterize the system behavior in a non-parametric manner, especially when the complexity of the system obstructs the establishment of a physics-based model. To take into consideration network traffic data and the sensor data, in \cite{bakalos2019protecting}, the authors proposed a multi-model data fusion and adaptive deep learning method based on a convolutional neural network to characterize the normal system behavior. The framework then detects cyberattacks as well as physical intrusions in a single ICS pertaining to water infrastructure. However, due to the complexity of the proposed method, it would not be applicable to a network of infrastructures, where a utility operates multiple equipment (subsystems) with interactions at the same time. Meanwhile, the establishment of such model requires thorough understanding of the system, and it is hard to generalize it to other systems.

\section{System Model}\label{sysmodel}
    \begin{figure*}[h!]
    \centering
	\includegraphics[scale=0.4]{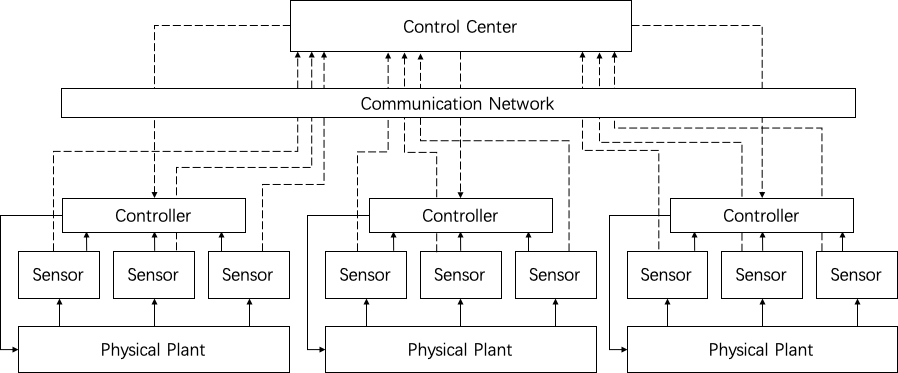}
	\caption{Networked Industrial Control System}
	\label{fig:system}
    \end{figure*}
We consider a networked control system consisting of $k$ subsystems as shown in Fig. \ref{fig:system}, where the subsystems are not necessarily homogeneous, meaning there is not a single model that can represent the behaviors of all the subsystems. For example, a smart grid is a network of generation buses, representing the power plants, and load buses, representing the substations, and different models should be used to represent the power plants and the substations. We denote the state of the subsystem $i$ at time $t$ by a vector $x^i_t$, where $x^i_t \in R^{n_i}$, and $n_i$ is the minimum number of variables needed to uniquely define the system state of site $i$. Denote the history of state of subsystem $i$ till time $t$ as $x^i_{\{t\}}=\{x^i_0,x^i_1,...,x^i_t\}$. The full system state vector at time $t$, $x_t$, is given by a concatenation of all the subsystem states. i.e, $x_t^T=[x^{1T},...,x^{kT}]\in R^{N}$ where $N=n_1+n_2+...+n_k$. Note that $x$ is not directly observed, but inferred from the measurements from the $M$ ($M>N$ for system observability) sensors distributed throughout the system. Similarly, we have $y_t^T=[y^{1T},...,y^{kT}]\in R^{M}$ where $M=m_1+m_2+...+m_k$, and $m_i$ is the number of sensors in subsystem $i$. Denote the control action as $u_t^T=[u^{1T},...,u^{kT}]\in R^{P}$, where $u^i_t\in R^{p_i}$ and $P=p_1+...+p_k$. Similarly, we denote the history of control actions and measurements till time $t$ as $u_{\{t\}}=\{u_0,u_1,...,u_t\}$ and $y_{\{t\}}=\{y_0,y_1,...,y_t\}$, respectively.

In general, the dynamics of the networked system can be represented by
\begin{gather}
    x_t=\mathcal{F}(x_{\{t-1\}},u_{\{t-1\}}),\label{ngs1}\\
y_t=\mathcal{G}_i(x_t),\label{ngs2}
\end{gather}
and a single subsystem $i$ can be represented by a recursive function of $x^i_t$ as follows:
\begin{gather}
    x^i_t=f_i(x_{\{t-1\}},u_{\{t-1\}})\label{gs1}\\
y^i_t=g_i(x_t)\label{gs2}
\end{gather}
Systems with linear dynamics where the Markovian property holds, can be represented by a linear state-space model. For an individual subsystem $i$ which operates independently, we have:
\begin{gather}
x^i_t=A_ix^i_{t-1}+B_iu^i_{t-1}\label{ss1}\\
y^i_t=C_ix^i_t\label{ss2}
\end{gather}

When the subsystems are correlated with each other, the full system can be modeled by
\begin{gather}
x_t=Ax_{t-1}+Bu_{t-1},\\
y_t=Cx_t,
\end{gather}
where $A$, $B$, and $C$ are functions of $A_i$'s, $B_i$'s, and $C_i$'s. In model-based methods, typically the above model is used to build a state estimator (e.g., a Kalman filter) that estimates $\hat{x}_t$ based on the observed measurement $y_t$ and the previous estimation $\hat{x}_{t-1}$. The estimated $\hat{x}$ is then substituted back into the system model to predict the next measurement $\hat{y}_{t+1}$, which is compared with the observed $y_{t+1}$ to obtain the residuals:
$$r_t=y_t-\hat{y}_t.$$

When accurately parameterized, the above model is robust and works well for simple systems. However, one disadvantage of these physics-based models is the estimation of all the parameters requires a large amount of data, and might cause the identifiability issues. Therefore, We assume the interconnectivity of these systems are complex to represent using a physics-based model, this is true especially when the system is nonlinear and the subsystems are interconnected.

Another disadvantage of the physics-based models is that sometimes they do not consider the dynamic behavior of the system or the temporal correlation in the data. In most systems, the control actions are calculated based on a period of historical data, or on a state estimation from the previous time step. In this case, the temporal correlation is implicit, and a steady-state estimation does not capture this type of correlation. For example, in a power system, the state vector $x$ contains the control actions $u_t$ as well.
%

We note that the control actions can be represented by the power generation setpoints. These setpoints can be determined by solving an operational planning problem subject to the system demand profile. Therefore, in this paper, we utilize the well known Mixed Integer Unit Commitment (MIUC) \cite{carrion2006computationally} as a means to compute our operational setpoints. The MIUC is widely used in the power industry for operational planning and can be formulated as follows \eqref{eq:centUC}:
\begin{subequations}\label{eq:centUC}
\begin{align}
\qquad & \underset{\alpha,\theta}{\text{min}} & c^\top \alpha+ d^\top \theta \label{eq:centUC0}&&\\
& \text{subject to} & Q\theta + R \alpha= E  &  \label{eq:centUC1}&\\ 
& &F \theta = H & \; \; \; & \label{eq:centUC2}
\end{align}
\end{subequations}

In Problem \eqref{eq:centUC} $\alpha$ represents a binary vector of length $|\mathcal{G}|\times T$ indicating if generators are turned on or turned off across each time epoch for each generator in the network, where $|\mathcal{G}|$ is the number of generators. Similarly, $\theta$ represents a real-valued vector of dispatch variables specifying the level of production on generators as well as the electric phase angles on separate buses of length $(|\mathcal{B}|+|\mathcal{G}|)\times T$, where $|\mathcal{B}|$ is the number of buses. 

Constraint \eqref{eq:centUC1} ensures that the commitment and production decisions are coupled along with ramp up (Q) and ramp down (R) constraints found in unit commitment. Constraint \eqref{eq:centUC2} enforces flow constraints (F) subject to the phase angle values as well as the transmission line capacities (H). 

As mentioned earlier, the implicit temporal correlation generated by calculating the control actions in a history-dependent manner cannot be easily represented by any physics-based model. On the other hand, the occurrence of an attack might be easily captured by analysing the temporal behavior of the system, while the steady state does not show any anomaly. Therefore, data driven method, specifically RNN, well fits this situation where temporal correlation needs to be captured via data-driven techniques.

\subsection{Covert Attack Model}
The covert cyberattack was first proposed in \cite{smith2011decoupled} for a single linear system as represented by functions \ref{ss1} and \ref{ss2}. Specifically, the attacker knows the values of $B_i$ and $C_i$. Recall that a covert attacker is assumed to possess access to control action and the sensor measurements as well. Under these assumptions, the attacker implements the covert attack by the following steps:
\begin{itemize}
    \item  First, the attacker manipulates the control actions using the following equation
    \begin{equation}
        \tilde{u}^i_t=u^i_t+a_t \label{ca1},
    \end{equation}
    where $\tilde{u}^i_t$ is the manipulated control action at time $t$, $a_t$ is the attack signal added to the original control action $u^i_t$. According to \ref{ss1}, this manipulation will alter the system state at $t+1$ by $B_ia_t$. That is,
    \begin{equation}
        \tilde{x}^i_{t+1}=x^i_t+B_ia_t 
    \end{equation}
    The consequent sensor measurements will be biased by $C_iB_ia_t$. That is,
    \begin{equation}
        \tilde{y}^i_{t+1}=y^i_{t+1}+C_iB_ia_t .
    \end{equation}
    \item Then, the attacker manipulates the sensor measurements by subtracting the above bias. i.e.,
    \begin{equation}
        \check{y}^i_{t+1}=\tilde{y}^i_{t+1}-\gamma,
    \end{equation}
    where $\gamma=C_iB_ia_t$. In this way, the manipulated measurement $\check{y}^i_{t+1}$ is equal to the expected measurement $y^i_{t+1}$ without attack. Hence, the covert attack can be successfully disguised, as the measurement is the only output of the system, which means all the data inference is conducted based on $y$.
\end{itemize}

In this work, we generalize the covert attack to nonlinear systems represented by Equations (\ref{gs1}) and (\ref{gs2}). We assume the attacker gains knowledge of the dynamics of subsystem $i$. This could be taken as the attacker obtains an estimation of the local functions $\hat{f}_i$ and $\hat{g}_i$ in Equations (\ref{gs1}-\ref{gs2}), which can serve as the simulator of system $i$. With this knowledge as well as the access to the control actions and sensor measurements, the attacker conducts a covert attack using the following steps:
\begin{itemize}
    \item  First, the attacker reads the original control action $u_t$ and simulates the expected sensor measurements using the knowledge of subsystem $i$. That is,
    \begin{gather}
        \hat{x}^i_{t+1}= \hat{f}_i(x_{\{t\}},u^i_{\{t\}}),\\
        \hat{y}^i_{t+1} = \hat{g}_i(\hat{x}^i_{t+1}).
    \end{gather}

    \item Then, the attacker manipulates the control actions as in Eq. (\ref{ca1}). According to Eq. (\ref{gs1}), this manipulation will alter the system state at $t+1$ as well as the sensor measurements. That is,
    \begin{gather}
        \tilde{x}^i_{t+1}=f_i(x_{\{t\}},\tilde{u}^i_{\{t\}},)\\
        \tilde{y}^i_{t+1} = g_i(\tilde{x}^i_{t+1}).
    \end{gather}

    \item Finally, the attacker replaces the consequent sensor measurements with the simulated one. i.e.,
    \begin{equation}
        \check{y}^i_{t+1} \leftarrow \hat{y}^i_{t+1}.
    \end{equation}
\end{itemize}
Notice that we assume the attacker's access to the sensors is limited to subsystem $i$. This means the attacker is not capable of compensating for the impact of attacking subsystem $i$ on other subsystems. This fact lays the foundation for our detection and localization framework. However, detecting the attack in this case is nontrivial because the sensors that are most informative of the attack are manipulated, while the attack's impact on other sensors does not have a clear indication of the occurrence of an attack as well as its location.

\section{Proposed Detection Framework}
    \begin{figure}[h!]
    \centering
	\includegraphics[scale=0.3]{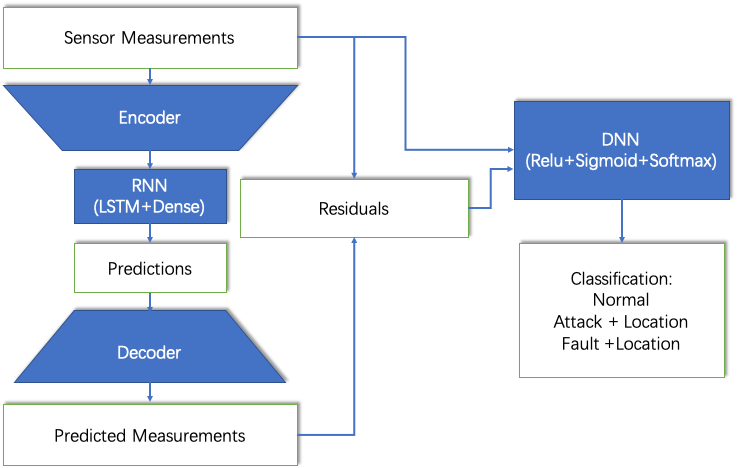}
	\caption{The proposed detection framework (AEN + RNN + DNN)}
	\label{fig:framework}
    \end{figure}
The structure of the proposed framework is shown in Fig.\ref{fig:framework}. The autoencoder is used for unsupervised feature extraction. Essentially, it projects the sensor measurements to a lower dimensional layer that filters out the noise and better represent the system status, which corresponds the steady-state state estimation in traditional frameworks. We use an RNN with an LSTM \cite{hochreiter1997long} layer to capture the nonlinear temporal dependency and the correlation among all the variables and extract the residuals. The RNN is used as a predictor, which corresponds to the particle filters or Kalman filters in the traditional frameworks. The residuals together with the original sensor measurements are fed into a DNN for detection as well as diagnosis, and the DNN corresponds to the detection and diagnosis schemes in the traditional frameworks.

Since the measurement $y\in R^m$ is obtained by the measurement function $\mathcal{G}(\cdot)$, which is a mapping from $x$ in the lower dimensional space $R^n$. The correlation among the sensor measurements is actually defined by the correlation among the state variables. Therefore, we use an autoencoder (AEN) to reduce the dimension of the sensor measurements, and for generality, the code size (output size of the encoder) is chosen as the dimension of the state vector, $n$. In our experiments, the autoencoder (and decoder) consists of 2 dense hidden layers with leaky ReLU activation functions. The encoded sensor measurements is taken as the input to the RNN (in our experiments, consisting of an LSTM layer followed by a dense layer). The output of the RNN is then decoded to reconstruct the predicted sensor measurements. The residuals are calculated as the difference between the observed measurements and the predicted ones. Then, we concatenate the residuals with the observed measurements, and the concatenated data are fed into a DNN. The DNN used in our experiments contains 3 dense hidden layers with ReLU, Sigmoid, and Softmax activation functions in sequence. The output of the DNN determines the type and location of the attack, designated by ``normal", ``attack+location", and ``fault+location". 

By the nature of covert attacks, the attacker has to obtain accurate system knowledge to conduct a successful attack. However, in reality, the complex subsystems are often geographically far apart from each other that makes it very unlikely that an attacker can obtain the knowledge necessary for attack, and get access to more than one subsystems. Therefore, in this work, we assume that the attacker only attacks one node, and it is less likely that faults on different nodes happen to occur at the same time. Hence, there are at most $2k+1$ independent possible conditions of the system, including "normal", \{``attack on node $i$, $i=1,...,k$"\}, and \{``fault on node $i$, $i=1,...,k$"\}.

The autoencoder and the LSTM are trained using only the data from normal operations. The DNN can be trained in a supervised manner using labeled data from simulation or historical data.
    
\section{Performance Evaluation}
\subsection{Data Extraction}
\begin{figure}[h!]
\centering
\includegraphics[scale=0.5]{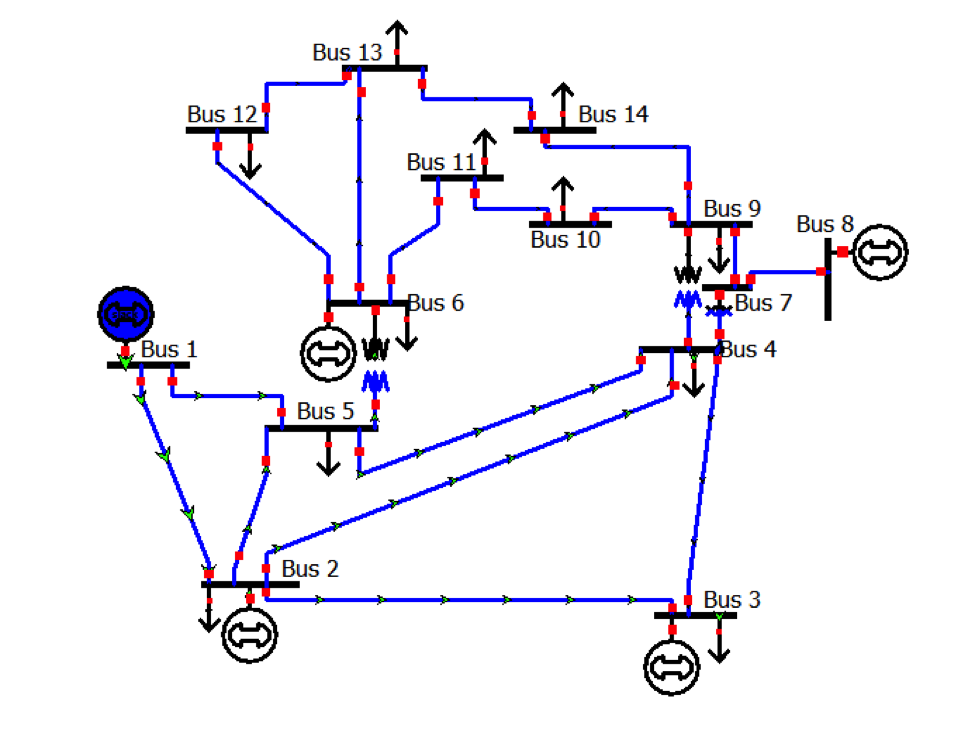}
\caption[The IEEE 14-Bus System]{The IEEE 14-Bus System\footnotemark}
\label{fig:14bus}
\end{figure}
\footnotetext{https://icseg.iti.illinois.edu/ieee-14-bus-system/}
We generate data via a simulation study on a smart grid. We use the IEEE 14-bus (Fig. \ref{fig:14bus}) to represent the smart grid at the transmission level to generate the time series of sensor measurements under different conditions. The model contains 9 load buses (buses 4, 5, 7, and 9-14), representing 9 substations; 4 generation buses (buses 2, 3, 6, and 8), representing 4 power generation plants; and 1 slack bus (bus 1), which is used to balance the active and reactive power in the system and also serves as a reference for all other buses. The model has 20 edges, representing the 20 transmission lines connecting the load and generator buses. The input to the simulation is the load profiles of all the load buses and the power generation plans of all the generation buses. We obtain the load profile of each substation by aggregation the load profiles of a random number of hourly residential power consumption profiles extracted from Pecan Street \cite{street2015dataport}. The generation plan is constructed using the method mentioned in Section \mbox{\ref{sysmodel}}. The simulation in Matlab uses Matpower \cite{zimmerman2010matpower} 7.0 to solve the power flow equation and add measurement noise. The output of the simulation is the hourly time series of the 39 sensor measurements for the simulated period. The 39 sensor measurements include the active power flow on each of the 20 transmission lines, the power generation of each generator bus as well as the slack bus, and the voltage of all the 14 buses.

Recall that the mechanism of a covert attack is to alter the system state by manipulating the control actions. Since in the transmission system we are considering in this simulation, most of the control happens in the power generation plants, we only consider the covert attacks on the generator buses. During the attack, the attacker decreases the generation level by a specific portion. The reason we choose to decrease the generation is from the attacker's objective: compared to generating more power than needed, decreasing the generation will cause possible blackouts and overloading of other generators, which is likely to cause more damage to the system. We simulate the covert attacks on each of the 4 generator buses at 5 levels of severity, where the attacker decreases the power generation by level 1: 20\%, level 2: 40\%, level 3: 60\%, level 4: 80\%, and level 5: 100\% of the planned generation. We assume the attacker obtains access to all the sensors related to the attacked generator and manipulates the sensor measurements by replacing the original values with the ones obtained from simulation such that it shows the attacked generator bus generates the same amount power as planned. For comparison, we simulate the faults as the decrease of power generation by the same amount caused by equipment malfunctioning. The biggest difference between a fault and a covert attack is there is no sensor data manipulation.

\subsection{Numerical Results}
\begin{figure}[htbp]
\centering
\includegraphics[scale=0.35]{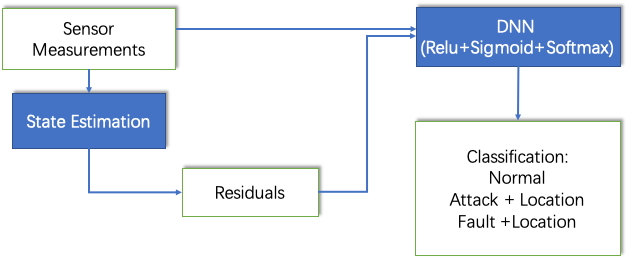}
\caption{The SE + DNN Structure}
\label{fig:sd}
\end{figure}

\begin{figure}[htbp]
\centering
\includegraphics[scale=0.35]{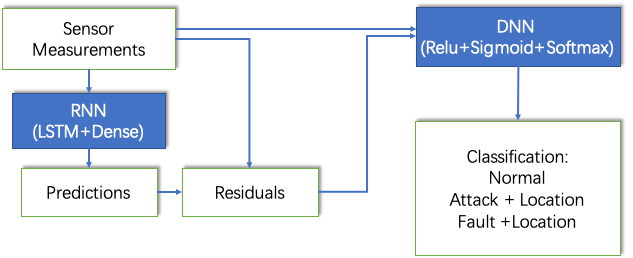}
\caption{The RNN + DNN Structure}
\label{fig:rd}
\end{figure}

We compare our proposed method with two benchmarks. The first one uses the traditional state estimation (SE) residuals for detection (Fig. \ref{fig:sd}). The state estimation is implemented by solving the (nonlinear) static power flow equations using Newton's method \cite{schweppe1970power}, which is implemented in Matpower 7.0. The second method uses the same RNN model as in the proposed model, but does not include the autoencoder (Fig. \ref{fig:rd}).

For the proposed model and the ``RNN + DNN" model mentioned above, we first calculate the first order difference of observations to ensure the stationarity, and then standardize the differences based on the standard deviation for each sensor. In this case, we use $n=14$ as the code size of the autoencoder. Usually for an AC $k$-bus system the state space has dimension $2(k-1)$, consisting of the magnitudes and phase angles of voltages at each bus. Since here we only care about the active power and do not consider the phase angle, we use $k-1=13$ as the code size. The stateful LSTM uses 10 lags to predict the next encoded observation. We train the autoencoder and the RNN with 80\% of the normal data, and keep them fixed when generating the residuals for attack and fault series. Therefore, the autoencoder and the RNN are only exposed to the normal data, which can be used to represent the system behavior under normal operations. The autoencoder and the RNN can be viewed as a non-parametric substitute of the physics-based models.

Since there are 39 sensors, the concatenation of residual and observation data is of length 78. Since there are 4 generator buses considered, the output of DNN is a multi-class classification of 9 labels listed in Tab \ref{tab:results}. The numbers refer to the bus number where the attack/fault occurs.

We train the model with 80\% of data and show the classification performance of the DNN by testing the model on the rest 20\% of data. The performance of the method is evaluated by precision, recall, and the F1-score, which are shown in Tab.\ref{tab:results} and Tab.\ref{tab:results2}, where Tab.\ref{tab:results} shows the classification performance of DNN among the 9 labels (normal, 4 attacks, and 4 faults), and Tab.\ref{tab:results2} shows the classification among the 3 classes (normal, attack, and fault) as well as the localization performance within the attack and fault classes. The precision, recall, and $F_1$-score are calculated based on the number of true positives (TP), true negatives (TN), false positives (FP), and false negatives (FN) using the following equations:
\begin{gather*}
    \text{Precision} = \frac{TP}{TP+FP}\\
    \text{Recall} = \frac{TP}{TP+FN}\\
    F_1 = 2\cdot\frac{\text{Precision}\cdot\text{Recall}}{\text{Precision}+\text{Recall}}
\end{gather*}
Tab.\ref{tab:results} and Tab.\ref{tab:results2} show that for most of the cases, the proposed method outperforms the other two methods in terms of detection as well as localization. In general, the SE+DNN model has similar but slightly better performance than the RNN+DNN model. However, both of them has a very low recall for attack on bus \#2. This is because in the network topology, bus \#2 has a high degree -- it has four neighbors, which is the highest among the four generator buses. According to the proposed generalization of covert attack in Section \ref{sysmodel}, the attacker manipulates all the sensors measuring the power flow on the four edges. Intuitively, in this case the attacker has the highest coverage of sensor accesses, which best covers the attack. Since the state estimation only refers to the data at current step and does not consider temporal correlation, it does not give a good estimation of the underlying truth. On the other hand, since the RNN is taking all the sensor measurements as input, the prediction of RNN is relatively inaccurate due to the noise in the sensor data. However, since the RNN considers the temporal correlation, the performance of RNN + DNN in this case is slightly better than SE + DNN. In contrast, the proposed method uses an autoencoder for unsupervised feature extraction, which helps filter out the noise in the data when training the RNN. In this case, the RNN gives more precise prediction of the data. Therefore, the residuals in this case could better capture the anomaly caused by the attack.

\begin{table*}[h!]
  \centering
  \caption{Precision, Recall, and F score for DNN classification}
    \begin{tabular}{c|ccc|ccc|ccc}
    \toprule
    Model & \multicolumn{3}{c|}{State Estimation + DNN} & \multicolumn{3}{c|}{RNN + DNN} & \multicolumn{3}{c}{Autoencoder + RNN + DNN} \\
    \midrule
    Label & Precision & Recall & F Score & Precision & Recall & F Score & Precision & Recall & F Score \\
    \midrule
    Normal     & 0.7330 & 0.9699 & 0.8350 & 0.7114 & 0.9803 & 0.8245 & 0.9980 & 0.9284 & 0.9619 \\
    Attack \#2     & 0.8761 & 0.1943 & 0.3180 & 0.8361 & 0.2203 & 0.3487 & 0.9998 & 0.9813 & 0.9904 \\
    Attack \#3     & 0.9985 & 0.9983 & 0.9984 & 0.9403 & 0.9921 & 0.9655 & 0.9773 & 0.9998 & 0.9884 \\
    Attack \#6     & 0.9998 & 0.9920 & 0.9959 & 0.9867 & 0.8146 & 0.8924 & 1.0000 & 0.8640 & 0.9271 \\
   Attack \#8     & 0.8645 & 0.9771 & 0.9173 & 0.9762 & 0.9046 & 0.9391 & 0.8918 & 0.9869 & 0.9369 \\
   Fault \#2     & 0.9999 & 1.0000 & 1.0000 & 0.9947 & 0.9995 & 0.9971 & 0.9999 & 0.9987 & 0.9993 \\
   Fault \#3    & 0.9925 & 0.9994 & 0.9959 & 0.9962 & 0.9774 & 0.9867 & 0.9997 & 0.9964 & 0.9981 \\
   Fault \#6    & 0.9308 & 0.9806 & 0.9551 & 0.8971 & 0.8792 & 0.8881 & 0.7320 & 0.9977 & 0.8444 \\
   Fault \#8    & 0.9572 & 0.9005 & 0.9280 & 0.9287 & 0.8930 & 0.9105 & 0.9729 & 0.8781 & 0.9231 \\

    \bottomrule
    \end{tabular}%
  \label{tab:results}%
\end{table*}%

\begin{table*}[h!]
  \centering
  \caption{Precision, Recall, and F score for detection and localization}
    \begin{tabular}{cc|ccc|ccc|ccc}
    \toprule
    \multicolumn{2}{c|}{Model} & \multicolumn{3}{c|}{State Estimation + DNN} & \multicolumn{3}{c|}{RNN + DNN} & \multicolumn{3}{c}{Autoencoder + RNN + DNN} \\
    \midrule
    \multicolumn{2}{c|}{Label} & Precision & Recall & F Score & Precision & Recall & F Score & Precision & Recall & F Score \\
    \midrule
    \multicolumn{2}{c|}{Normal} & 0.7330 & 0.9699 & 0.8350 & 0.7114 & 0.9803 & 0.8245 & 0.9980 & 0.9284 & 0.9619 \\
    \multicolumn{2}{c|}{Attack} & 0.9778 & 0.7199 & 0.8292 & 0.9512 & 0.6817 & 0.7942 & 0.9900 & 0.9667 & 0.9782 \\
    \multicolumn{2}{c|}{Fault} & 0.9776 & 0.9885 & 0.9830 & 0.9683 & 0.9565 & 0.9624 & 0.9180 & 0.9896 & 0.9524 \\
    \midrule
    \multirow{4}[2]{*}{Attack} & \#2   & 0.9962 & 0.9468 & 0.9709 & 0.9685 & 0.9914 & 0.9798 & 0.9999 & 0.9815 & 0.9906 \\
          & \#3   & 0.9999 & 1.0000 & 1.0000 & 0.9913 & 1.0000 & 0.9956 & 0.9819 & 0.9999 & 0.9908 \\
          & \#6   & 0.9999 & 1.0000 & 1.0000 & 0.9991 & 0.9869 & 0.9929 & 1.0000 & 1.0000 & 1.0000 \\
          & \#8   & 0.9460 & 0.9961 & 0.9704 & 0.9951 & 0.9565 & 0.9754 & 1.0000 & 1.0000 & 1.0000 \\
    \midrule
    \multirow{4}[2]{*}{Fault} & \#2   & 0.9999 & 1.0000 & 1.0000 & 0.9998 & 0.9999 & 0.9998 & 1.0000 & 1.0000 & 1.0000 \\
          & \#3   & 1.0000 & 0.9999 & 1.0000 & 0.9996 & 0.9998 & 0.9997 & 1.0000 & 0.9999 & 1.0000 \\
          & \#6   & 1.0000 & 1.0000 & 1.0000 & 0.9998 & 0.9992 & 0.9995 & 1.0000 & 1.0000 & 1.0000 \\
          & \#8   & 1.0000 & 1.0000 & 1.0000 & 0.9973 & 0.9980 & 0.9977 & 0.9997 & 1.0000 & 0.9998 \\
    \bottomrule
    \end{tabular}%
  \label{tab:results2}%
\end{table*}%

\begin{figure*}[htbp]
\centering
\includegraphics[scale=0.6]{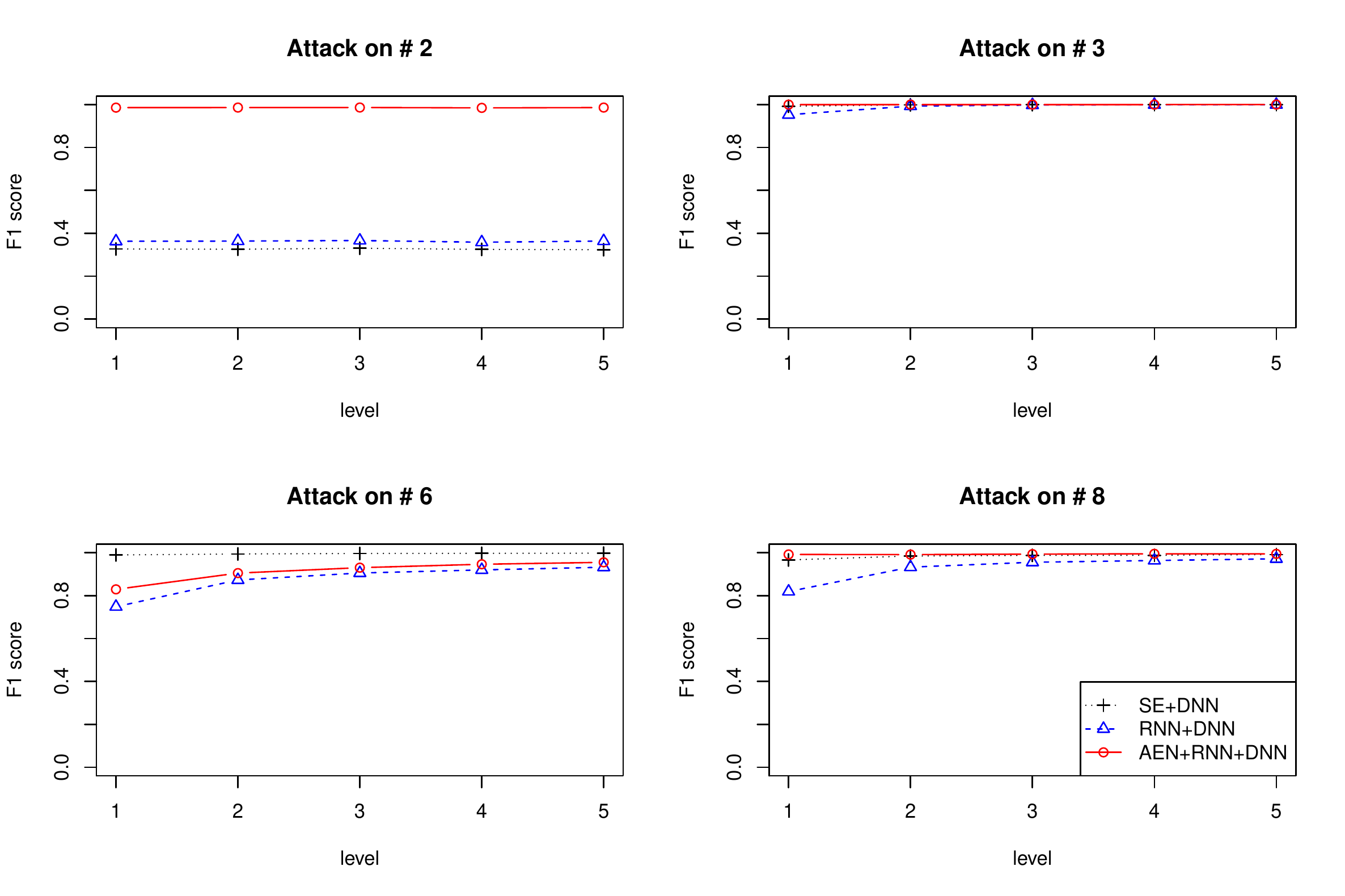}
\caption{F1 score under different levels of attack for the three models}
\label{fig:F1}
\end{figure*}

We also show the $F_1$ scores of the three methods under different levels of attacks and faults in Fig.\ref{fig:F1}. In general, the proposed method has a better performance than the other two, especially for attacks on bus \#2. Another inspection is that the $F_1$ score increases as the level (severity) of the attack increases. This is a validation that the covertness (the ability to stay undetected) of the attack decreases as the the severity of attack increases, meaning the distinction between normal data and the data under attack becomes clearer as the severity of attack increases, leading to a higher detection power and diagnosis accuracy. Moreover, it can be seen that the performance of all three methods generally depends on the connectivity of the attacked node -- Attacks on bus \#2 and \#6 has lower detection rates because they have more neighbors.

\begin{figure}[h!]
\centering
\includegraphics[scale=0.5]{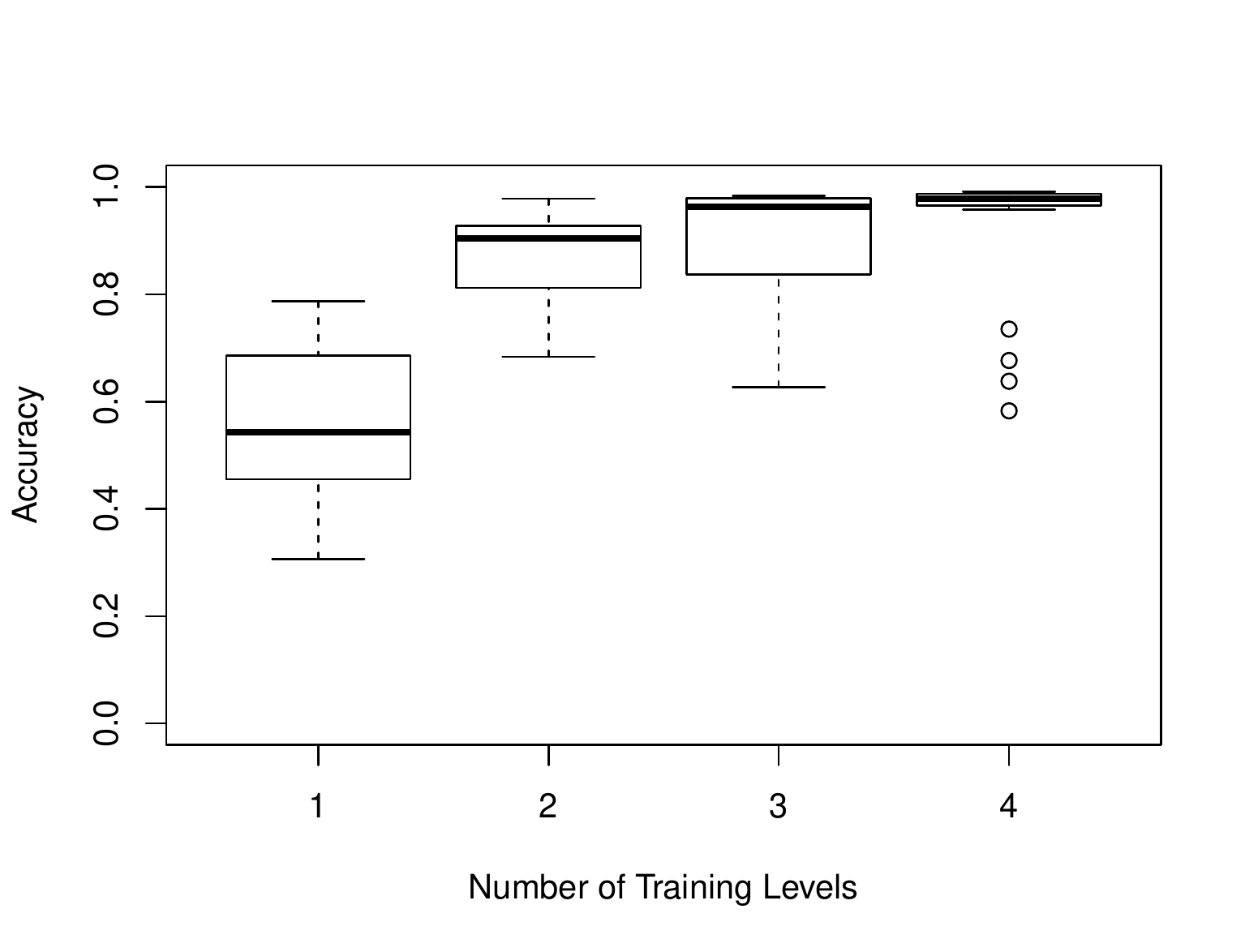}
\caption{The boxplots of accuracy under different number of training levels}
\label{fig:box}
\end{figure}

To evaluate the performance of the proposed method when encountering attacks that are not in training, we use a subset of the attack levels to train the DNN and test it on the other levels. We test the method by selecting $l$ ($l=1,2,3,4$) levels among the five levels simulated. For each $l$, we replicate the training and testing 20 times. Within each replication, the attack levels for training is randomly selected. The boxplot of testing accuracy is shown in Fig. \ref{fig:box}. The result shows that the proposed method has a relatively high accuracy when the model is trained on more than 2 levels of attacks. Meanwhile, as more attack levels are used in training the model, the testing accuracy increases, and the variance of the accuracy decreases. When the testing levels are out of the range of the training levels, the performance is worse (e.g., the outliers for $l=4$ are corresponding to the replications where the model is trained with levels 2-5 and tested on level 1). This is because if the data is trained on a stronger attack, the data from a weaker attack would lie in between the clusters of normal data and the attack data, and hence the DNN would have difficulty classifying these data.

\section{Conclusion}
In this paper, we proposed a generic data-driven framework for detecting, diagnosing, and localizing covert attacks on industrial control systems. The proposed framework uses an autoencoder for unsupervised feature extraction from sensor measurements, and then uses an RNN to capture the temporal correlations among the encoded sensor measurements. The prediction of the RNN is decoded and compared with the input sensor measurements to get the residuals, which help detect the anomaly. The residuals and the sensor measurements are processed with a DNN to determine whether an observation is representing normal conditions, an attack, or a fault, and to identify the location of the attack/fault. The proposed framework was compared with the model-based state estimation technique, as well as a modification of itself by removing the autoencoder. The results showed that the autoencoder helps extract the important features from the data, as well as reduce the dimension of the input to the RNN. This significantly helps improve the classification accuracy of the DNN. It is worth noticing that the RNN does not provide a more accurate estimation of the state compared to the model-based state estimation. However, since the RNN considers the temporal behavior of the system, which is not considered by the mode-based SE, the residuals obtained from the decoded RNN prediction could better capture the anomalous characteristics of the data when the system is under the attacks/faults. The reason model-based SE does not perform well under covert attack is because the objective of SE is to minimize the residuals. This leverages the estimation of the state to normal conditions, which does not represent the underlying truth, especially when the attacker has access to more sensors. The simulation study and performance evaluation validated the proposed method. Since this method is model-free, it is easily generalizable to other networked industrial control systems.

In this work, we only trained and tested the model on the known attack/fault types. A future direction is to extend the method to anomaly-based detection, which can detect novel attacks and faults. Another direction is to combine the method with graphical network topology, as well as the correlation structure of the data.

\bibliographystyle{elsarticle-num}    
\bibliography{ref1} 

\end{document}